\documentclass[journal]{IEEEtran}
\usepackage{cite}
\usepackage{amsmath,amssymb}
\usepackage{amsthm}
\usepackage{nccmath}
\usepackage{fancyhdr}
\usepackage{setspace}
\usepackage{svg}
\usepackage{bbm}
\usepackage{soul}
\usepackage{subfloat}
\usepackage{graphicx}
\usepackage{textcomp}
\usepackage{xcolor}
\usepackage{kotex}
\usepackage{subfigure}
\usepackage{booktabs}
\usepackage{soul,color}
\usepackage{bbold}
\usepackage{mwe}
\usepackage{multicol}
\usepackage{multirow}
\usepackage[normalem]{ulem}
\usepackage[linesnumbered,ruled]{algorithm2e}
\usepackage{cite}
\usepackage{amsmath,amssymb,amsfonts}
\usepackage{bbm}
\usepackage{graphicx}
\usepackage{textcomp}
\usepackage{xcolor}
\usepackage{kotex}
\usepackage{algorithmicx,algpseudocode}
\usepackage{setspace}
\usepackage{amsthm,thmtools,thm-restate}
\usepackage{subfigure}
\usepackage{booktabs}
\usepackage[linesnumbered,ruled]{algorithm2e}

\newtheorem{theorem}{Theorem}

\begin{document}

\title{Two Tales of Platoon Intelligence for Autonomous Mobility Control: Enabling Deep Learning Recipes}

\author{
    Soohyun Park, 
    Haemin Lee, 
    Chanyoung Park, 
    Soyi Jung, 
    Minseok Choi, and \\
    Joongheon Kim,~\IEEEmembership{Senior Member, IEEE}  
    \thanks{Soohyun Park, Haemin Lee, Chanyoung Park, and Joongheon Kim are with the Department of Electrical and Computer Engineering, Korea University, Seoul 02841, Republic of Korea (e-mails: \{soohyun828,haemin2,cosdeneb,joongheon\}@korea.ac.kr).}
    \thanks{Soyi Jung is with the Department of Electrical and Computer Engineering, Ajou University, Suwon 16499, Republic of Korea (e-mail: sjung@ajou.ac.kr).}
    \thanks{Minseok Choi is with the Department of Electronic Engineering, Kyung Hee University, Yong-in 17104, Republic of Korea (e-mail: choims@khu.ac.kr).}
    \thanks{Joongheon Kim is a corresponding author.}
}
\maketitle
\begin{abstract}
This paper presents the deep learning-based recent achievements to resolve the problem of autonomous mobility control and efficient resource management of autonomous vehicles and UAVs, i.e., (i) multi-agent reinforcement learning (MARL), and (ii) neural Myerson auction.
Representatively, communication network (CommNet), which is one of the most popular MARL algorithms, is introduced to enable multiple agents to take actions in a distributed manner for their shared goals by training all agents' states and actions in a single neural network. 
Moreover, the neural Myerson auction guarantees trustfulness among multiple agents as well as achieves the optimal revenue of highly dynamic systems. 
Therefore, we survey the recent studies on autonomous mobility control based on MARL and neural Myerson auction. Furthermore, we emphasize that integration of MARL and neural Myerson auction is expected to be critical for efficient and trustful autonomous mobility services.\end{abstract}

\IEEEpeerreviewmaketitle

\section{Introduction}

In the fast-paced world of technological advancements, autonomous mobility has emerged as a transformative innovation, dramatically reshaping numerous aspects of human life, such as transportation, logistics, and surveillance~\cite{obj_jung}. These complex systems depend on advanced algorithms, sensors, and communication networks to carry out their tasks smoothly and proficiently with their own objectives~\cite{survey}. One crucial element that supports the successful functioning of these systems, particularly when operating as a coordinated group, is the efficient sharing of information among multiple mobility platforms. 

The information exchange is essential for the seamless operation of a networked platoon, whether it consists of autonomous vehicles navigating on roadways or drones flying in coordinated patterns. This data sharing facilitates collaboration among individual units, empowering them to maintain their formation, prevent collisions, and execute synchronized maneuvers~\cite{platoon-01}. As a result, an effective information exchange system considerably bolsters the overall safety, dependability, and performance of these autonomous fleets. In addition to ensuring safe and efficient operations, information exchange in platoon systems also facilitates resource optimization, energy conservation, and reduced environmental impact. By sharing critical data, such as velocity, position, and route information, platoons can synchronize their movements and optimize their trajectories, ultimately resulting in decreased fuel consumption, emissions, and traffic congestion. This not only contributes to a more sustainable future but also helps organizations achieve their economic and environmental objectives.

In this paper, we investigate two deep learning methods for addressing the challenges associated with managing and controlling autonomous mobility platoons, i.e., reinforcement learning (RL) and neural Myerson auction. Reinforcement Learning, a prevalent technique in artificial intelligence, enables autonomous agents to learn from their environment in order to accomplish specific objectives via the maximization of their own expected returns. When applied to autonomous mobility applications, RL can be used to derive optimal control strategies for maintaining safety, efficiency, and robustness in various traffic situations. Furthermore, in order to control the platoon, the use of single-agent RL is not suitable because all agents will identically operate when they are located in a same space and time with same action-reward settings. Therefore, for realizing the cooperation and coordination among multiple agents, multi-agent RL (MARL) algorithms should be utilized~\cite{RL-01, RL-02, RL-03}. Among various MARL algorithms, this paper considers communication network (CommNet) which is widely and actively used in modern distributed computing and networking applications. With CommNet, the centralized neural network training with multi-agents' states and actions is conducted at first, and then, the distributed execution is realized by sharing this centralized trained neural networks with multiple agents. Therefore, this type of MARL algorithm is named to \textit{centralized training and distributed execution (CTDE)}. 

Conversely, neural Myerson auctions, an emerging approach in the field of mechanism design, combines the power of deep learning with the traditional Myerson auction framework. Among traditional action mechanisms, second price auction (SPA) is widely used for truthful trading/bidding of auctioneer's setting items under the consideration of truthfulness of participating bidders. However, the major problem of SPA is the non-optimality of auctioneer's revenue. Therefore, in order to improve the revenue, the Myerson auction is utilized with the concept of virtual valuation. Furthermore, according to the recent advances in deep learning methodologies, the Myerson auction can be re-designed, and the algorithm is now called \textit{neural Myerson auction}. This method can effectively allocate resources among multiple autonomous vehicles while preserving incentive compatibility, leading to efficient and fair outcomes, and furthermore, truthfulness is preserved. 

By integrating these two major deep learning approaches, i.e., (i) CTDE-based MARL and (ii) neural Myerson auction, designing a comprehensive solution for managing and controlling autonomous mobility platoons, addressing various challenges ranging from vehicle cooperation and coordination to resource allocation among multiple autonomous vehicles. 

The major contributions of this paper can be summarized as follows.
\begin{itemize}
    \item First of all, the major research trends for the use of CTDE-based MARL algorithms are discussed in various autonomous mobility control platoon applications. For more details, refer to Sec.~\ref{sec:2}.
    \item Moreover, the neural Myerson auction is introduced and explained which is mainly used for multi-user resource allocation and scheduling under the consideration of truthfulness among multiple users. In addition, the major applications of neural Myerson auction are well explained. For more details, refer to Sec.~\ref{sec:3}. 
\end{itemize}

The rest of this paper is organized as follows.
Sec.~\ref{sec:2} and Sec.~\ref{sec:3} introduce the main concepts and applications of CTDE-based MARL and neural Myerson auction for various autonomous mobility control applications. Sec.~\ref{sec:4} concludes this paper.

\section{CTDE-based MARL for Autonomous Mobility}\label{sec:2}

\subsection{Legacy Technologies}
The cooperative and coordinated autonomous mobility control has been studied with various approaches. First of all, convex optimization and linear optimization based approaches are widely used~\cite{titsconvex}. The advantage of this approach is that optimality can be preserved. However, it has significant drawbacks, i.e., \textit{(i)} all information for optimization formulation and computation should be provided in each computation procedure which is not realistic, and \textit{(ii)} the information sharing for the optimization formulation and computation should be conducted without delays. Furthermore, Lyapunov control-based approach can be also considered which is for time-average utility maximization subject to stability~\cite{ton201608kim,jsac201806choi,tmc201907koo}. According to the fact that this approach is fully distributed, it is beneficial for multiple autonomous mobility control platforms. However, unexpected environmental dynamics cannot be considered with this Lyapunov control-based algorithms. Therefore, it is essentially required to consider learning-based approaches.

\subsection{MARL Basics}
\begin{figure*}[ht!]
  \centering
  \includegraphics[width=0.99\linewidth]{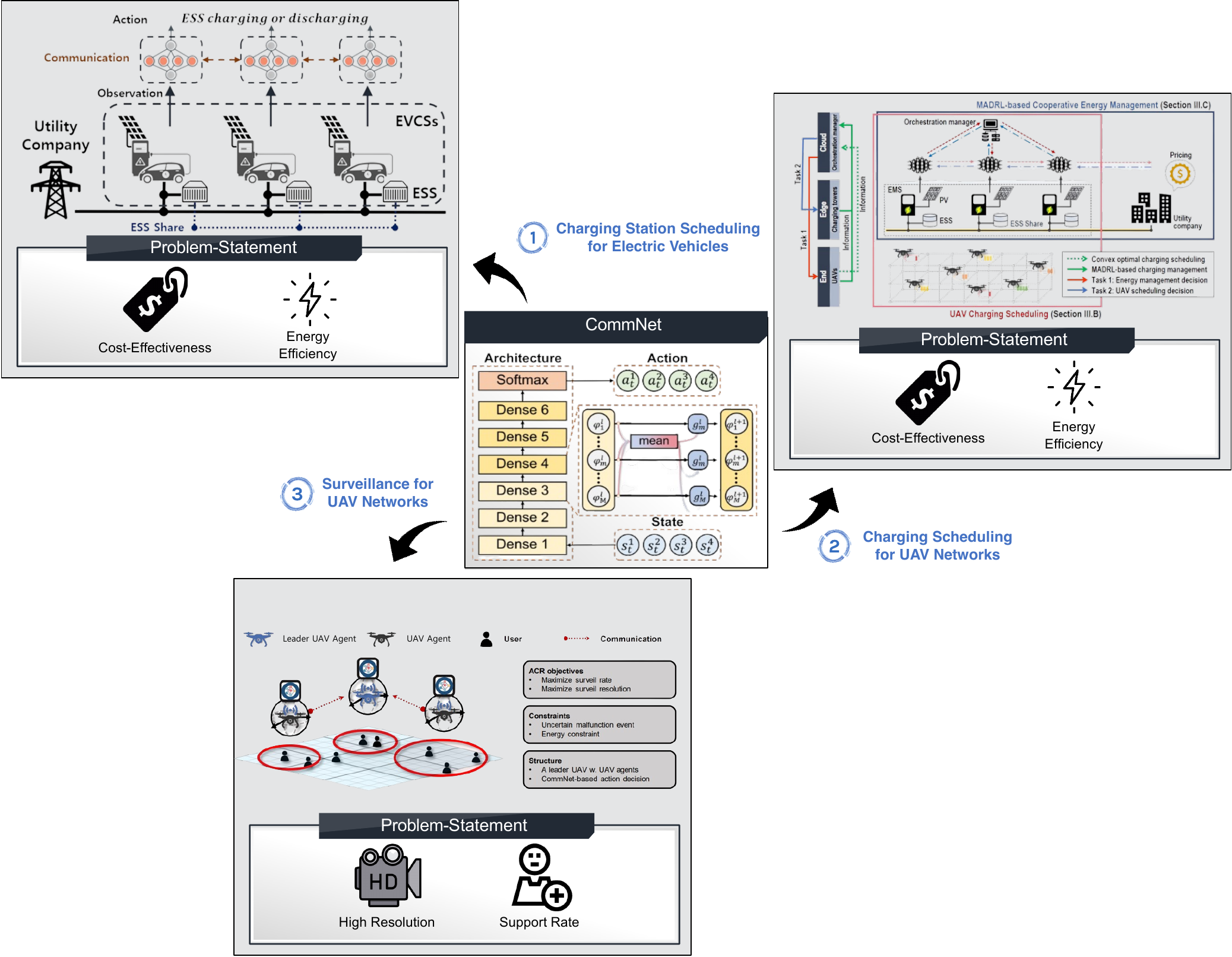}
  \caption{Representative papers utilizing CommNet for the cooperation of multiple agents in multifaceted MARL scenarios.}
  \label{fig:sysmodel}
\end{figure*}

Reinforcement learning (RL) is a sequential decision-making strategy in dynamic environments~\cite{mnih2013playing}. RL has shown superior performances in controlling autonomous mobility~\cite{grigorescu2020survey} since agents trained with RL can adaptably react to environmental uncertainties.
However, in multi-agent reinforcement learning (MARL), multiple agents can have an impact on other agents' policy training~\cite{iotj2023yun,iotj2010kwon,iotj23park,aimlab2022icte,aaai23quantum}. This non-stationarity resulting from the interaction of multiple agents makes conventional RL-based approaches struggle to accomplish optimal policy~\cite{tan2022cooperative}.
In order to overcome this challenge, many researchers have focused on creating relevant MARL algorithms that can break through the limitations of single-agent RL by enabling agents to interact and coordinate with one another.
One novel approach to find an optimal policy in MARL is using a communication network (CommNet)~\cite{sukhbaatar2016learning}. CommNet is an advanced neural network architecture that allows agents to find optimal policies to better achieve common objectives by allowing agents to share information observed in the environment with each other when training policies. Recent works have demonstrated the MARL supremacy of CommNet in versatile scenarios for operating autonomous mobility as presented in Fig.~\ref{fig:sysmodel}.

\subsection{CommNet}

\begin{figure*}[t]
    \centering
    \includegraphics[width=\linewidth]{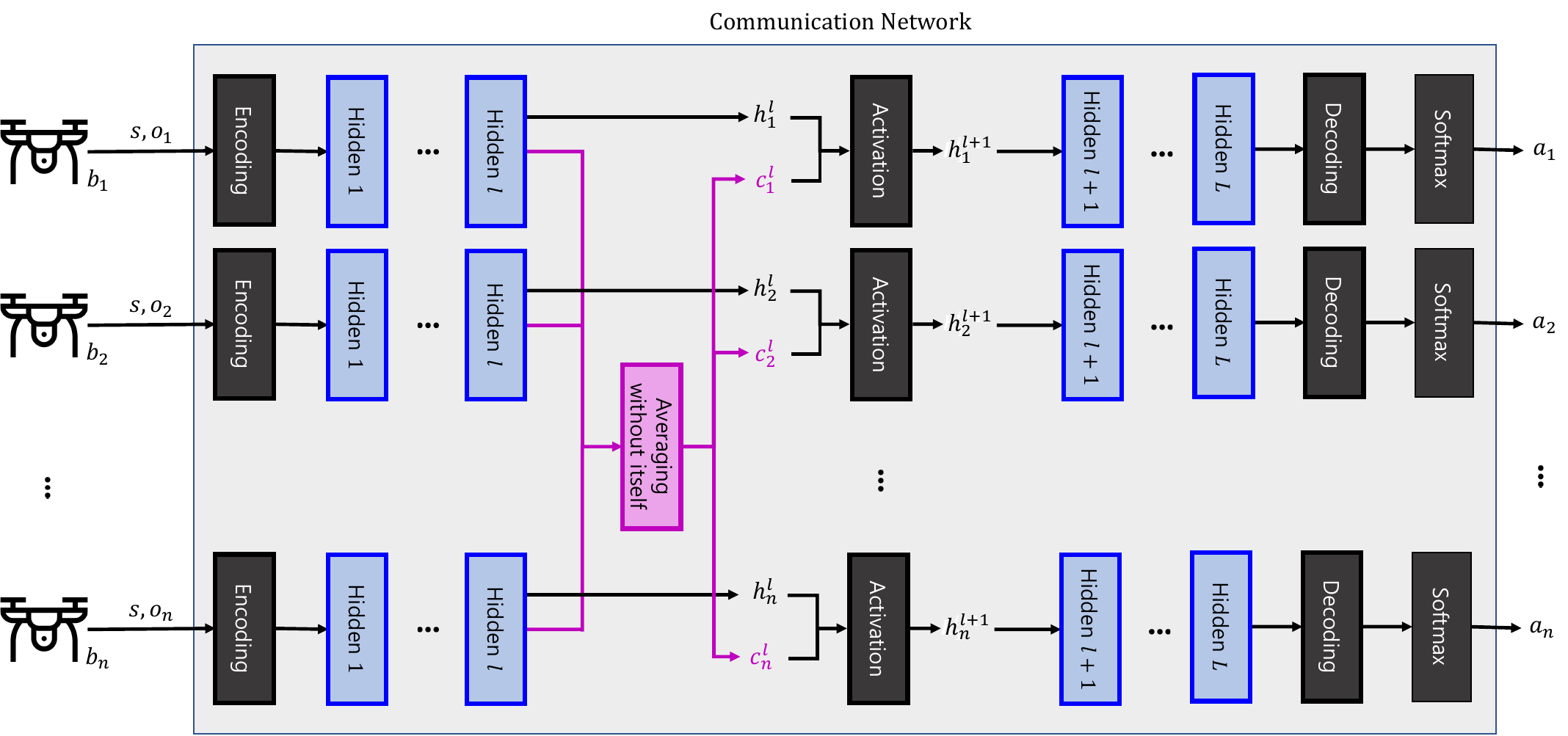}
    \caption{CommNet-based multi-agent reinforcement learning framework.}
    \label{fig:CommNet}
\end{figure*}

As mentioned, CommNet is utilized for multiple agents to cooperatively achieve a common objective through inter-agent communications as illustrated in Fig.~\ref{fig:CommNet}. Basically in MARL, all agents represented in $b_1,b_2,\cdots b_n$ get their experiences by getting a partial observation $o$ of the environment. In addition, the ground truth state $s$ is decided by the sequential decision-making of all agents. These state and observations are feed-forwarded into the neural network of each agent as input. After encoding the input vector, the hidden variable $h^l$ goes through several $L$-dense layers which are fully connected with the next hidden layers. At every step moving to the next dense layer, the CommNet-based agent creates a communication variable $c^l$ by averaging hidden variables of other agents except itself to share information between agents. This communication step allows multiple agents to learn their neural network parameters cooperatively. Before the hidden variable and communication variable are entered into the next layer, they are concatenated and activated with a non-linear function such as ReLU, hyperbolic tangent, or Sigmoid function. When this process is repeated to reach the last layer, the action probabilities are returned by decoding the hidden variable of the last dens layer and taking the softmax function. After the policy training process, multiple agents can cooperatively achieve a shared goal in a distributed manner with their trained policies without centralized control or explicit coordination rules.

\subsection{Applications}

For the applications of CommNet, cooperative and coordinated energy/power scheduling algorithms for multiple platforms are considered because energy-related discussion is the most important in power-hungry battery-operated mobility platforms. Therefore, charging scheduling algorithms for battery-operated mobility platforms are discussed in Sec.~\ref{sec:2-3-1} and Sec.~\ref{sec:2-3-2}. Lastly, surveillance is one of the most important tasks in multi-UAV networks. Therefore, the research results for CommNet-based multi-UAV surveillance are presented in Sec.~\ref{sec:2-3-3}.

\subsubsection{Charging Station Scheduling for Electric Vehicles}\label{sec:2-3-1}
In the Industry 4.0 Revolution era, the demand for electric vehicles (EVs) is growing exponentially. EVs can control the motor rotation without latency in an eco-friendly way, ensuring that they are more suitable for autonomous vehicles than conventional vehicles with an internal combustion engine. Furthermore, electronic components of EVs have an advantage over internal combustion engines in analyzing vehicle data and diagnosing faults through low current and voltage. For supporting an increasing number of EVs, a cost-efficient energy management system of EV charging stations (EVCSs) is required to be designed based on supplier-consumer behavior patterns. Existing optimization strategies in a centralized manner are challenging to deal with high system dynamics and massive data in real time. Therefore, Shin \textit{et al.} proposed a CommNet-based optimization approach to manage a vast amount of data in a distributed manner with consideration of 
the state of an energy storage system (ESS) and photovoltaic (PV) power production for EVCSs~\cite{shin2019cooperative}. In their scenario, every EVCS can serve energy to EVs while charging energy with its PV charger. An EVCS only shares surplus energy with other EVCSs after meeting its own demand in order to reduce overall operating costs by managing surplus energy. Via data-intensive performance evaluations, it is confirmed that every CommNet-based EVCS agent can jointly minimize the amount of purchasing energy from the private enterprise while observing the energy state and prices fluctuating throughout the day.

\subsubsection{Charging Tower Scheduling for UAV Networks}\label{sec:2-3-2}
Thanks to unmanned aerial vehicles (UAVs)' high agility and mobility, they can take various and flexible roles in a wide range of fields. However, an efficient energy management system is also vital in UAVs due to their limited battery capacity~\cite{9447255}. Thus, Jung \textit{et al.} proposed an optimization framework for joint charging scheduling and CommNet-based energy management assisted by the cloud~\cite{jung2021orchestrated}. In their scenario, multiple charging towers provide charging services to any UAVs during run-time operations in a decentralized manner, based on the plug-and-play method. All charging towers taking roles of independent RL agents train their policies in the direction of efficiently operating UAV networks at low cost. They cooperatively share their energy to minimize the overall network operation cost which stands for the joint objective of agents. This work also presented the convex-optimal charging scheduling scheme for the fairness among UAVs in terms of resource allocations.

\subsubsection{Surveillance for UAV Networks}\label{sec:2-3-3}
UAVs are also employed to support the autonomous mobile surveillance system~\cite{zhang20196g}. Different from conventional CCTVs that are static, UAVs can provide on-demand surveillance by dynamically updating the locations. Moreover, they can monitor extreme environments where ground mobility cannot access owing to physical limitations. Nevertheless, UAVs still have to overcome environmental uncertainties, \textit{i.e.,} collision with obstacles, or battery fully discharging, and manage coverage regions to avoid their overlapping for service efficiency. Therefore, it is important to carefully operate and control UAVs to provide reliable and flexible autonomous surveillance services. To deal with this problem, Yun \textit{et al.} proposed a CommNet-based multi-UAV positioning in which there are a leader UAV and multiple non-leader UAVs~\cite{yun2022cooperative}. The leader UAV makes decisions based on the observation of all UAVs, whereas non-leader UAVs take only on their own observation. In this paper, UAVs monitor a large number of users with high video resolution by adaptively moving two-dimensional trajectories or controlling the video resolution. Therefore, the authors of \cite{yun2022cooperative} balance the tradeoff between surveillance areas and video resolution, and also investigate the impacts of malfunctioned UAVs.

\section{Distributed Learning for Autonomous Mobility}\label{sec:3}

\subsection{Legacy Technologies}
The distributed resource allocation and scheduling in autonomous mobility control has been actively studied. First of all, convex optimization-based approaches are also considerable by constructing optimization formulations for resource allocation~\cite{auction-ref1,auction-ref2,auction-ref3}. As presented, the advantage of this approach is that optimality is guaranteed. However, distributed operations are not possible because all information for optimization formulation and computation should be provided. In addition, game theory based approaches can be useful for this application, which can solve distributed optimization under uncertainty~\cite{game-1,game-2}. However, game theory formulation cannot consider truthfulness which can be preserved with our considering auction-based approaches. Lastly, Lyapunov control-based approach can be also considered due to the nature of distributed computation~\cite{ton201608kim,jsac201806choi,tmc201907koo}. However, as mentioned, unexpected environmental dynamics cannot be considered for the computation.

\subsection{Auction Basics}
A conventional first-price auction (FPA) is one of the well-known types of auction procedures; A bidder who submits the highest bid value to \textit{auctioneer} (which is also called \textit{seller}) is awarded and pays its bid value to the auctioneer. Consider a scenario where there are $N$ bidders, i.e., 
\begin{equation}
b_{1},\cdots, b_{N}, 
\end{equation}
and an auctioneer, and bidders' bid values are 
\begin{equation}
v_{1},\cdots, v_{N}. 
\end{equation}
The auctioneer selects the maximum bid value, which is denoted as $v^{*}$, with \begin{equation}
v^{*}=\max\{v_{1},\cdots, v_{N}\}; 
\end{equation}
and the winner bidder $b^{*}$ is the one who submitted bid value $v^{*}$. Suppose that the second highest bid value is $v^{\dag}$. The winner bidder $b^{*}$ does not need to pay the entire bid amount of $v^{*}$, as submitting a bid slightly higher than $v^{\dag}$ is sufficient to win the auction. As a result, individual bidders must be strategic when participating in an FPA. However, the presence of untruthful bidders undermines the incentive compatibility of the mechanism, making FPA inefficient~\cite{fpa}.

Another type of auction mechanism is called second price auction (SPA), which functions similarly to FPA in terms of selecting a winner. However, instead of paying their highest bid value, the winner pays the second highest bid value. SPA is widely regarded as a truthful auction mechanism in the literature \cite{tvt201905shin,infocom2017} and is commonly used for resource allocation in distributed computing applications \cite{srcho1,srcho2}. However, SPA has a drawback in that it cannot achieve revenue optimality, as the auctioneer only receives the second-highest bid value instead of the highest. 
In order to address this issue, various approaches have been studied. Among them, Myerson auction with the concept of virtual valuation is one of the well-known approaches \cite{tvt201905shin, auction_tii}. Monotonic increasing functions are commonly used to numerically formulate virtual valuation. With the advancements in deep neural network (DNN) research, the Myerson auction computation procedure can be approximated using DNNs. As a result, this paper proposes a DNN-based autonomous aerial delivery scheduling algorithm, called the neural-architectural Myerson auction, which uses a DNN to compute the virtual valuation.

\subsection{Neural Myerson Auction for Optimal Delivery}\label{sec:auction}

This section discusses the use of deep learning-based auction to maximizes the expected revenue of surveillance drone while guaranteeing truthfulness and revenue-optimal. The monotonic network is used for random sampling to approximate pseudo-optimal revenue values. Additionally, allocation networks and payment networks are used to determine the winner drone and the payment, respectively. The detailed neural architectures for deep learning to solve the proposed auction-based problems are presented in the following subsections, i.e., monotonic networks (refer to Sec.~\ref{sec:4-1}), allocation networks (refer to Sec.~\ref{sec:4-2}), and payment networks (refer to Sec.~\ref{sec:4-3}).

\begin{figure}\centering
    \includegraphics[width=3in]{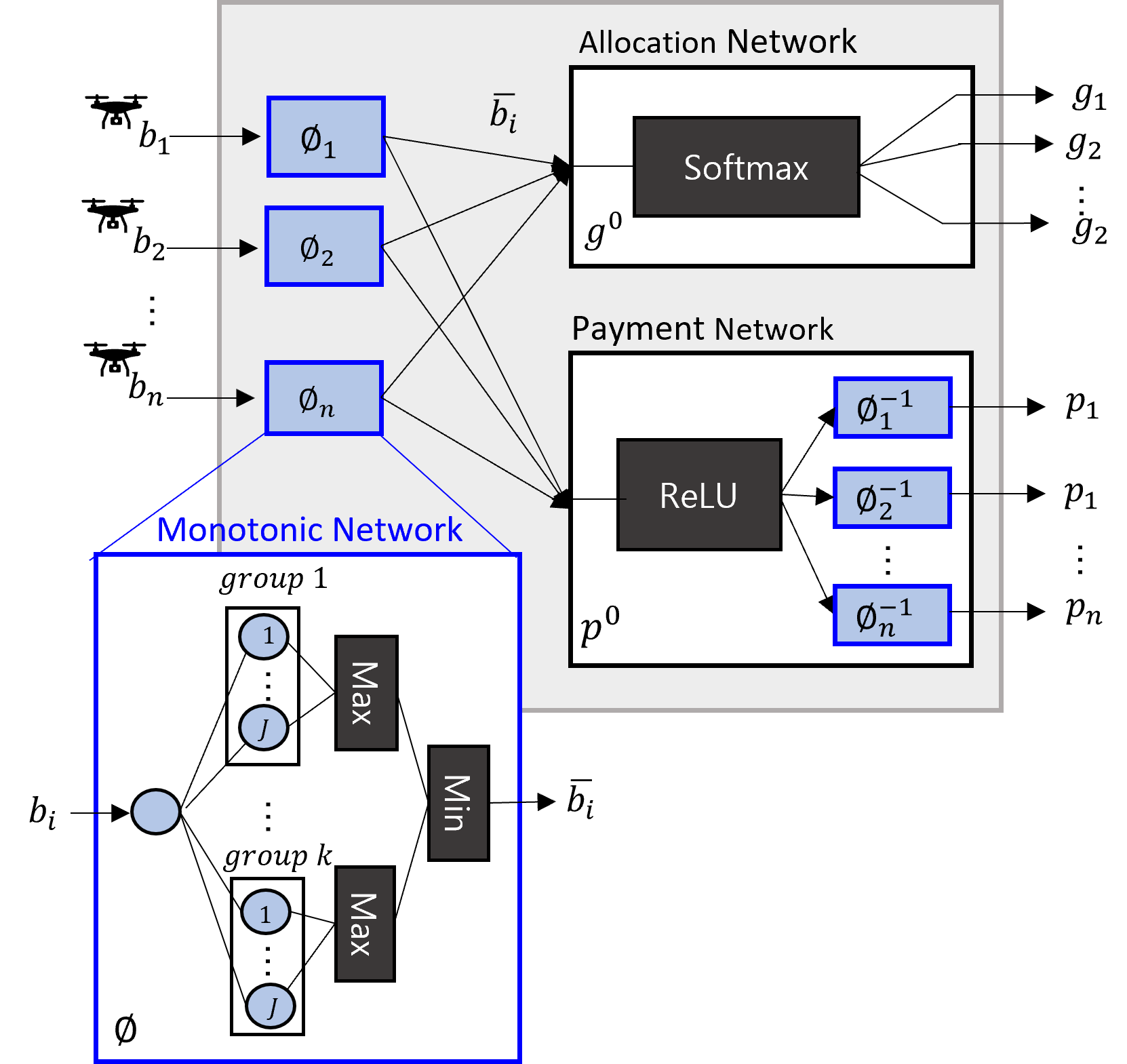}
    \caption{Deep learning auction framework}
    \label{fig:framework}
\end{figure}

\subsubsection{Virtual Valuation Function}\label{sec:4-1}
The virtual valuation function in auction network is represented by $\phi_i$ and serves as the virtual valuation in Myerson auction~\cite{luong2018optimal}. The input bids $b_i$ of delivery drones are transformed into $\bar{b_i}$ after passing through the monotonic network, which performs max/min operations over several linear functions. The Monotonic network $\phi_i$ is composed of  $K$ groups of $J$ linear functions and is defined as follows~\cite{luong2018optimal}.

\subsubsection{Winner Determination Function}\label{sec:4-2}
The SPA allocation network maps the surveillance drone and delivery drone with the highest non-zero transform bid. The output of the network is the allocation probabilities, which are determined by a softmax layer using the transformed bids $\bar{b_i}$ and a dummy input $\bar{b_{N+1}} = 0$. Semantically, the softmax layer is used to determine the maximum value. The allocation network using softmax can be expressed as follows,

\begin{eqnarray}
g_i(\bar{b}) &=& \textsf{softmax}_i(\bar{b}_{1},...,\bar{b}_{N+1} ; k) \\
&=& \frac{e^{k\bar{b}_i}}{\sum_{j=1}^{N+1}e^{k\bar{b}_j}}, \forall i \in N
\label{eq:3}
\end{eqnarray}
where $k$ is a parameter of softmax function and it determines the quality of the approximation~\cite{tvt201905shin,luong2018optimal}.

\subsubsection{Payment Function}\label{sec:4-3}

The payment network utilizes a rectified linear unit (ReLU) activation function to ensure that the final payment made to the winning delivery drone is non-negative. The ReLU activation function can be represented as follows,

\begin{equation}
p_i^{0}(\bar{b}) = ReLU(\max_{\forall j \neq i}\bar{b_j}), \forall i \in N.
\label{eq:4}
\end{equation}

Finally, the final payment made by the winner delivery drone to surveillance drone can be calculated as follows,
\begin{equation}
p_i = \phi_i^{-1}\left(p^{0}_i(\bar{b})\right).
\end{equation}
\subsubsection{Neural Network Training and Complexity}\label{sec:4-4}

Neural architecture trains parameters, $w^{i}_{kj}$ and $\beta^{i}_{kj}$ are trained using a training set of valuation profiles and a loss function that minimizes the negative revenue in Myerson auction. The loss function $\hat{R}$ is defined as follows,

\begin{equation}
\hat{R}(w,\beta) =-\sum_{i=1}^{N}  g_i^{(w,\beta)}(v^{s})p_i^{(w, \beta)}(v^{s}).
\label{eq:5}
\end{equation}

The results of allocation networks and payment networks are used for training parameters, and stochastic gradient descent optimizer is employed to  the loss function $\hat{R}$.

In conventional deep learning computation procedures, two phases exist, i.e., (i) \textit{training} and (ii) \textit{inference}. During the training phase, it takes time for training for cost function minimization with iterative computation such as stochastic gradient descent for backward propagation. Most work evaluates the complexity as a training time. The training time was around 5 minutes running on the CPU (Intel i7, 8 cores) and RAM (16GB). On the other hand, during the inference phase, conducting simple dense layer computation with trained optimal/approximated parameters is required which are the matrix computation and activation function computation. Therefore, the computation time consists of a monotonic network computation with several layers (i.e., the algorithm complexity can be linearly scaled). It can be represented as 
\begin{equation}
(O_{M}(m)+O_{A}(m)) \times NL, 
\end{equation}
where $O_{M}(m)$ and $O_{A}(m)$ is the computation complexity of the matrix operation for each layer~\cite{zhu2020revenue}. Here, $m$ and $NL$ denotes node number and number of layers. After the training, the real-time execution in the inference phase can be done within a seconds.


\subsection{Auction Properties}
 In the preceding section, we define the truthful characteristics and the auction network including the allocation rule $g$ and payment rule $p$. According to Myerson theorem, the truthful condition IR and IC can be ensured.

\begin{theorem}[Myerson~\cite{myerson1981optimal}]
For single parameter environments, any set of strictly monotone functions $\phi_1, \phi_2,...,\phi_N$, an auction that assigns an item to the bidder with highest virtual valuation $\phi_i(v_i)$ and the payment is determined by the second highest virtual valuation is IR and IC.
\end{theorem}

The neural architecture consists of $K$ groups with outputs labeled as $t_1, t_2,...,t_R$. Within each group, the number of hyperplanes is denoted by $h_r$, where $r$ ranges from $1$ to $R$. The hyperplane parameters are represented by 
\begin{equation}
\mathbf{w}{(r,1)}, \mathbf{w}{(r,2)}, \cdots,\mathbf{w}{(r,h_r)},     
\end{equation}
and the entire set of weights and biases is given by the matrix $\mathbf{W}$. Group $r$ produces an output, i.e.,  
\begin{equation}
t_r(x)=\displaystyle\min{j}(\mathbf{w}{(r,j)}\cdot \mathbf{x} + \theta{(r,j)}), 1 \le j \le h_r, 
\end{equation}
and the final output is as,
\begin{equation}
\mathbf{O_x} = \displaystyle\max_{r}{\mathit{t_r(x)}}. 
\end{equation}

This is the same as the virtual valuation network in Sec.~\ref{sec:4-1}. If all weights in the first layer are positive, this network satisfies increasing monotonicity, which is the condition satisfied in our system~\cite{daniels2010monotone}.

\subsection{Applications}
\subsubsection{Data Delivery in UAV Networks}
Several studies have explored data acquisition frameworks for wireless sensor networks using drones to improve the efficiency of data collection. S. Say, \textit{et. al.}~\cite{say2016priority} proposed a priority-based frame selection approach to reduce redundant data transmissions between sensor nodes and drones. Algorithm in~\cite{zhan2017energy, data_tii} leverages drones as mobile data collectors for randomly deployed sensor nodes. This study aims to jointly optimize the wake-up schedules of sensor nodes and the trajectories of drones to minimize the maximum energy consumption, thereby achieving min-max fairness. J. Gong, \textit{et. al.}~\cite{gong2018flight} proposed an algorithm that minimizes the total flight time of drones while ensuring that each sensor can upload a certain amount of data. Additionally, C. Singhal, \textit{et. al.}~\cite{singhal2021aerial}  proposed an adaptive surveillance and event-telecast video streaming service using drones that employ WiFi-direct link scheduling and dynamic configuration settings to transmit data to ground control stations.

In this paper, we present an aerial data delivery scheduling approach that utilizes a neural Myerson auction computation. This approach is different from previous studies in which delivery drones compete to directly transfer data. Our approach has an advantage over previous studies in that it can enable data transmission sustainably in extremely poor conditions. Additionally, our deep learning-based auction reduces costs by selecting the optimal delivery drone for the data collection process. Although our study is consistent with other research on drone-based data delivery networks that consider the energy and coverage of drones as bid values, it distinguishes itself by enabling data delivery even in the absence or destruction of communication infrastructure.

\subsubsection{Resource Allocation in UAV Networks}\label{sec:application}
The use of auction-based approaches is a popular and effective method for solving resource allocation and scheduling problems in a distributed and truthful manner. However, there is inherent uncertainty involved in the valuation process for both the auctioneer/seller and the buyers/bidders. The auctioneer is uncertain about the true values that bidders attach to the object being sold, which is the maximum amount each bidder is willing to pay. If the auctioneer had complete information about the values, they could simply offer the object to the bidder with the highest value at or below what the bidder is willing to pay. However, bidders do not know the true values attached by other bidders, and knowing these values would not affect how much the object is worth to them~\cite{krishna2009auction}. Due to the large volume of economic transactions that are conducted through auctions, auction-based computation processes have been extensively studied for limited resource allocation and scheduling problems~\cite{klemperer1999auction}. 
B. Dai, \textit{et. al.}~\cite{dai2014price} examined a transportation problem that aims to reduce costs and increase profits by collaborating carriers with pick-up and delivery requests. They introduced a pricing-setting combinatorial auction to solve the problem. In contrast, D. C. Marinescu, \textit{et. al.}~\cite{marinescu2013auction} proposed a self-organizing architecture for large-scale compute clouds and a scalable combinatorial auction-based solution for cloud resource allocation. On the other hand, B. Coltin, \textit{et. al.}~\cite{coltin2013online} presented an auction-based scheduling algorithm to enhance the efficiency of item deliveries between robots. The algorithm works online and adjusts to new requests, dead vehicles, and shared information. In addition, auction-based incentive mechanism for collaborative computation offloading that can achieve near-optimal long-term social welfare is proposed in~\cite{truthful_tii}.

Our considering Myerson auction is one of the most efficient revenue-optimal single-item auctions~\cite{myerson1981optimal}, among various auction-based scheduling and resource allocation algorithms. To numerically approximate the Myerson auction, a DNN-based architecture can be used, which has led to the development of learning-based Myerson auction algorithms for charging scheduling in wireless power transfer (WPT)-based multi-drone networks and electric vehicles, as proposed in~\cite{tvt201905shin} and~\cite{lee2021truthful}, respectively. In addition, DNN-based auctions have been used to solve resource allocation problems in mobile edge computing and wireless virtualization, as demonstrated in~\cite{zhu2020revenue} and~\cite{luong2018optimal}. Moreover, the proposed algorithm in~\cite{kuo2020proportionnet} aims to address concerns of fairness while maintaining high revenue and strong incentive guarantees by approximating auctions using deep learning.

\section{Concluding Remarks}\label{sec:4} 
This paper investigates the challenges of autonomous mobility in a multi-agent system with high system dynamics. 
First, we focus on a CTDE-based MARL algorithm that allows autonomous entities to pursue the shared objective and make decisions in a distributed manner based on their separate observation and investigate several application studies based on CommNet which is a popular CTDE-based MARL algorithm.
However, a MARL-assisted decision method has difficulty in guaranteeing all agents' trustfulness and achieving revenue optimality; therefore, we secondly present the neural Myerson auction. 
The neural Myerson auction efficiently allocates system resources for distributed autonomous entities, e.g., vehicles and UAVs, while achieving high revenue and strong incentives. 
Thus, we conclude that the integrated framework of MARL and neural Myerson auction is powerful for efficiently controlling multiple agents with autonomous mobility and pursuing the trustful system revenue at the same time.

\bibliographystyle{IEEEtran}
\bibliography{ref_aimlab}

\end{document}